\documentclass{article}
\usepackage{graphicx} 
\usepackage{amsmath}
\usepackage{array}
\usepackage{multirow}
\usepackage{amssymb}
\usepackage{subfig}
\usepackage{algorithm}
\usepackage{algpseudocode}
\usepackage{enumerate}
\usepackage[numbers]{natbib}

    \usepackage[preprint]{neurips_2025}

\usepackage[utf8]{inputenc} % allow utf-8 input
\usepackage[T1]{fontenc}    % use 8-bit T1 fonts
\usepackage{hyperref}       % hyperlinks
\usepackage{url}            % simple URL typesetting
\usepackage{booktabs}       % professional-quality tables
\usepackage{amsfonts}       % blackboard math symbols
\usepackage{nicefrac}       % compact symbols for 1/2, etc.
\usepackage{microtype}      % microtypography
\usepackage{xcolor}         % colors

\title{RoHyDR: Robust Hybrid Diffusion Recovery for Incomplete Multimodal Emotion Recognition}

\author{
  Yuehan Jin \\
  School of Computer Science and Engineering\\
  South China University of Technology\\
  \texttt{jinyuehan2004@gmail.com} \\
  \And
  Xiaoqing Liu \\
  School of Future Technology\\
  South China University of Technology\\
  Pengcheng Laboratory\\
  \texttt{ft\_liuxiaoqing@mail.scut.edu.cn} \\
  \AND
  Yiyuan Yang \\
  Department of Computer Science\\
  University of Oxford\\
  \texttt{yiyuan.yang@cs.ox.ac.uk} \\
  \And
  Zhiwen Yu \\
  School of Computer Science and Engineering\\
  South China University of Technology\\
  \texttt{zhwyu@scut.edu.cn} \\
  \AND
  Tong Zhang \\
  School of Computer Science and Engineering\\
  South China University of Technology\\
  \texttt{tony@scut.edu.cn} 
  \And
  Kaixiang Yang \\
  School of Computer Science and Engineering\\
  South China University of Technology\\
  \texttt{yangkx@scut.edu.cn} 
}

\begin{document}

\maketitle

\begin{abstract}
Multimodal emotion recognition analyzes emotions by combining data from multiple sources. However, real-world noise or sensor failures often cause missing or corrupted data, creating the Incomplete Multimodal Emotion Recognition (IMER) challenge. In this paper, we propose Robust Hybrid Diffusion Recovery (RoHyDR), a novel framework that performs missing-modality recovery at unimodal, multimodal, feature, and semantic levels.
For unimodal representation recovery of missing modalities, RoHyDR exploits a diffusion-based generator to generate distribution-consistent and semantically aligned representations from Gaussian noise, using available modalities as conditioning. For multimodal fusion recovery, we introduce adversarial learning to produce a realistic fused multimodal representation and recover missing semantic content. We further propose a multi-stage optimization strategy that enhances training stability and efficiency.
In contrast to previous work, the hybrid diffusion and adversarial learning-based recovery mechanism in RoHyDR allows recovery of missing information in both unimodal representation and multimodal fusion, at both feature and semantic levels, effectively mitigating performance degradation caused by suboptimal optimization. Comprehensive experiments conducted on two widely used multimodal emotion recognition benchmarks demonstrate that our proposed method outperforms state-of-the-art IMER methods, achieving robust recognition performance under various missing-modality scenarios. Our code will be made publicly available upon acceptance.

\end{abstract}

\section{Introduction}

Multimodal Emotion Recognition (MER) is a fundamental task in affective computing, which aims to understand human emotions by jointly analyzing heterogeneous modalities, including textual, acoustic, and visual signals. MER has demonstrated significant potential across various domains, including marketing~\cite{6,7}, social media analysis~\cite{8,9}, robotics~\cite{2,3}, and human-computer interaction~\cite{4,5}. While most existing studies assume complete availability of all modalities~\cite{11,Mai33}, this assumption rarely holds in real-world applications~\cite{MTMSA12}. In dynamic environments, modality data may be partially missing due to various technical and environmental constraints~\cite{GCNet10}. This challenge gives rise to the task of Incomplete Multimodal Emotion Recognition (IMER), which focuses on emotion recognition under incomplete modality conditions. Given its practical significance and inherent challenges~\cite{GCNet10, IMDer13}, IMER has garnered increasing research attention. Existing approaches can be broadly categorized into two main streams: representation learning-based approaches and generation model-based approaches.

\begin{figure}[!t]
% \vspace{-2.0em}
  \centering
  \includegraphics[width=1\textwidth]{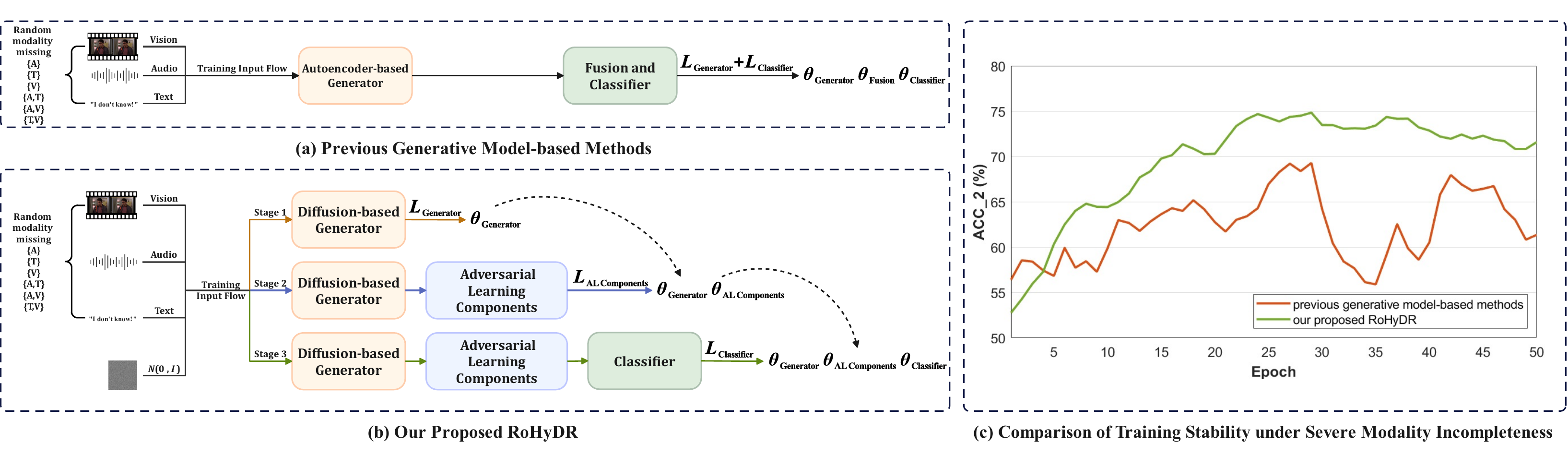} 
  \caption{(a) Previous generative model-based methods typically employ a single-level autoencoder-based generator to recover each missing unimodal input.  
(b) Our proposed Robust Hybrid Diffusion Recovery (RoHyDR) integrates a diffusion-based generator for unimodal recovery and adversarial learning components for multimodal recovery. Additionally, the multi-stage optimization strategy effectively mitigates the issue of suboptimal convergence.  
(c) RoHyDR exhibits more stable training dynamics across 50 training epochs.}
  \label{fig 1}
  % \vspace{-2.0em}
\end{figure}

Representation learning-based methods aim to extract a joint semantic representation from available modalities to infer the semantics of missing ones~\cite{MTMSA12, MMIN14, TATE15, SMCMSA16, NIAT20}. However, these approaches often neglect modality-specific feature distributions that are crucial for maintaining each modality's discriminative power~\cite{IMDer13}, resulting in semantic and structural deviations in the recovered information.
Generative model-based methods employ generators to recover missing modality features that align with the target distribution~\cite{DiCMoR17, LDM18, DDPM21}, as illustrated in Figure~\ref{fig 1}(a). However, the inherent uncertainty in generative models frequently leads to semantic mismatches between generated and real features~\cite{24,25,26}. These discrepancies affect multimodal fusion, producing joint representations that deviate from those obtained with complete modality inputs.
Overall, current research methods either exclusively focus on enhancing unimodal reconstruction capability or solely consider maintaining 
multimodal semantic consistency in the joint representation, resulting in suboptimal reconstruction of missing modalities and limited recovery in multimodal representation robustness. On the other hand, the significant disparity in loss values between internal modules such as the generator and classifier continuously creates a gradient imbalance, degrading training stability and efficiency, as shown in Figure~\ref{fig 1}(c).

Motivated by the challenges outlined above, we tackle the IMER problem by introducing a novel method, Robust Hybrid Diffusion Recovery (RoHyDR), as shown in Figure~\ref{fig 1}(b). In detail, to overcome the limitations of existing approaches in recovering missing modalities, RoHyDR adopts a hybrid two-module architecture: an unimodal representation recovery module and a multimodal fusion recovery module. These are designed to recover missing information at the unimodal and semantic levels, in both unimodal representation and multimodal fusion. The \emph{unimodal representation recovery} module incorporates a generator consisting of a high-dimensional diffusion model and a unimodal reconstructor to generate semantically aligned and distribution-consistent representations. It leverages Gaussian noise to generate missing modalities, using the available modalities as conditioning signals to enhance semantic correlation. The \emph{multimodal fusion recovery} module adopts adversarial learning~\cite{GAN19}, incorporating a fusion network, a multimodal reconstructor, and a discriminator. This module aims to generate a realistic fused multimodal representation and recover missing semantic content. The discriminator learns to distinguish between fused representations from complete inputs and those recovered from the unimodal representation recovery module. This adversarial setup encourages the fusion network and multimodal reconstructor to produce a robust representation that closely resemble those from complete modality inputs, effectively compensating for semantic loss. Finally, the recovered multimodal representation is fed into a classifier for emotion prediction. To address the imbalance in training objectives across modules, RoHyDR adopts a multi-stage optimization strategy. This strategy leads to more stable training, as illustrated in Figure~\ref{fig 1}(c).

The main contributions of this work are threefold. (1) We propose RoHyDR, a novel framework for incomplete multimodal emotion recognition that performs unimodal representation recovery using Gaussian noise through a diffusion-based generator and incorporates multimodal adversarial learning through a discriminator-driven multimodal fusion recovery mechanism. Both recoveries are on the feature-level and the semantic-level. (2) A multi-stage optimization strategy is developed to effectively address the challenges in multi-objective function optimization, enabling systematic optimization of missing modality recovery, multimodal joint representation learning, and robust emotion classification. (3) Extensive experiments on two multimodal emotion recognition benchmarks demonstrate that RoHyDR achieves state-of-the-art performance or competitive results under various missing modality scenarios.

\section{The Proposed RoHyDR Method}
\begin{figure}[!t]
% \vspace{-2.0em}
  \centering
  \includegraphics[width=1\textwidth]{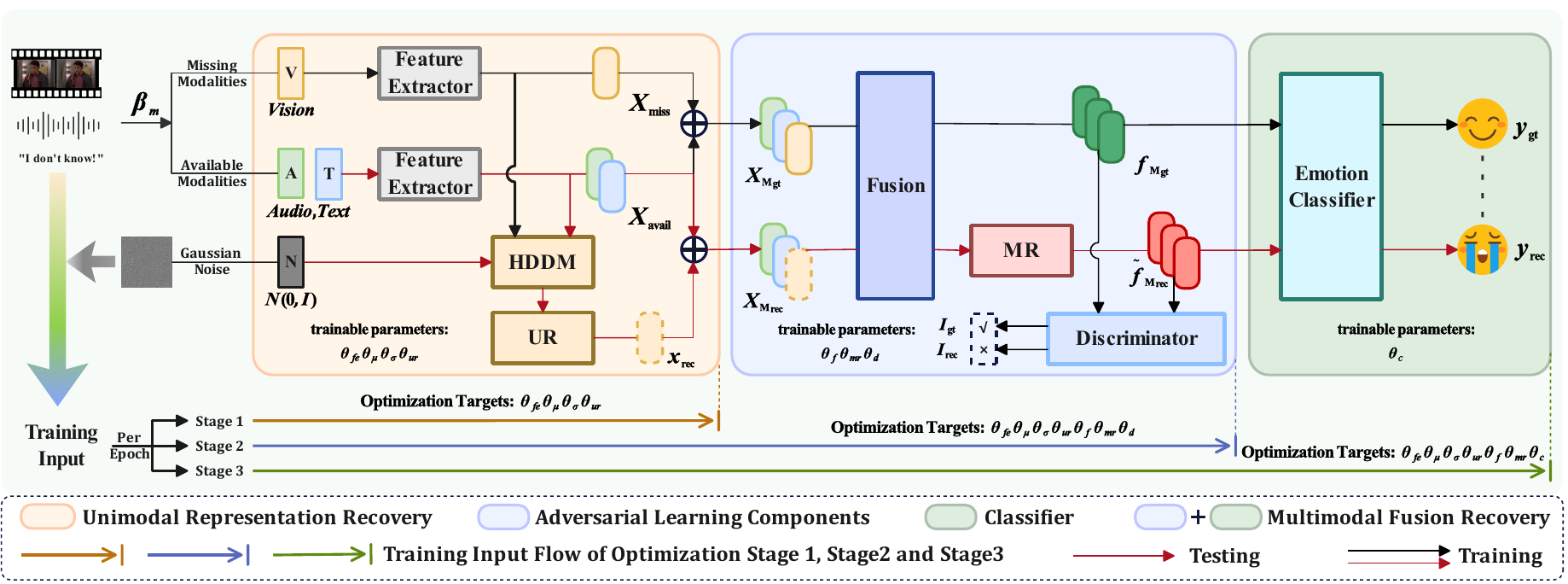} % 替换为你的图片路径
  \caption{The framework of the proposed RoHyDR. In the context of IMER, multimodal data \([A; T; V]\) is categorized into two types: available modalities and missing modalities. Any single modality or any pair of modalities may be missing (vision modality is missed as an example here). During training, RoHyDR receives the ground-truth of the missing modalities, available modalities, and Gaussian noise as inputs. During testing, only the available modalities and Gaussian noise are provided as inputs.}
  \label{fig 2} % 可选，用于交叉引用
  % \vspace{-1em}
\end{figure}
\subsection{Problem Definition and Proposed Framework}
Figure~\ref{fig 2} illustrates the overall framework. In the MER task, each sample is represented as a tuple \([A; T; V]\), where \(A\), \(T\), and \(V\) denote the input of audio, text, and vision, respectively. To model modality incompleteness, we design a binary missing indicator \(\beta_m \in \{0,1\}\) for each modality \(m \in \{a, t, v\}\), where \(\beta_m = 0\) indicates the availability of modality \(m\), and \(\beta_m = 1\) indicates its absence. We represent the ground truth of missing representation as \(X_{\text{miss}} = [\beta_a \cdot x_a;\, \beta_t \cdot x_t;\, \beta_v \cdot x_v]\), where \(x_m\) is the representation of modality \(m\). The available modalities are defined as \(X_{\text{avail}}\).

The proposed hybrid recovery strategy is detailed in Section~\ref{Hybrid Recovery Method}. 
During training, the \emph{unimodal representation recovery} module (Section ~\ref{Unimodal Representation Recovery}) first extracts modality-specific representation, including the missing ones \(X_{\text{miss}}\) and the available ones \(X_{\text{avail}}\). Secondly, a high-dimensional diffusion model (HDDM), conditioned on \(X_{\text{avail}}\) and Gaussian noise, collaborates with a unimodal reconstructor (UR) to generate the recovered representation \(x_{\text{rec}}\), to approximate \(X_{\text{miss}}\). Subsequently, the \emph{multimodal fusion recovery} module (Section~\ref{Multimodal Fusion Recovery}) employs a discriminator to distinguish between the ground truth fused representation \(f_{\text{M}_{\text{gt}}}\) and the recovered counterpart \(\tilde{f}_{\text{M}_{\text{rec}}}\), producing feedback signals \(I_{\text{gt}}\) and \(I_{\text{rec}}\). These signals guide both the fusion network and the multimodal reconstructor (MR) to generate a more robust fused representation. The classifier is optimized using the predicted labels \(y_\text{gt}\) and \(y_\text{rec}\), derived respectively from \(f_{\text{M}_{\text{gt}}}\) and \(\tilde{f}_{\text{M}_{\text{rec}}}\).
During testing, the model operates solely on the available modalities \(X_{\text{avail}}\) and Gaussian noise, without access to the missing modality ground truth.

The multi-stage optimization strategy is presented in Section~\ref{Multi-Stage Optimization Strategy}, consisting of three sequential stages that progressively optimize unimodal recovery, multimodal fusion, and final emotion classification. 
Stage 1 aims to enhance the model's capability for unimodal recovery by using only the generator loss with the first input flow. Stage 2 improves the robustness of the multimodal fused representation from incomplete modalities by using only adversarial loss with the second input flow.
Stage 3 adapts the model for the final emotion recognition objective.

\subsection{Hybrid Recovery Method}
\label{Hybrid Recovery Method}

\subsubsection{Unimodal Representation Recovery}
\label{Unimodal Representation Recovery}
Each modality, including audio, text, and vision, is processed by a dedicated feature extractor \(\mathrm{FE}_{\theta_{fe}}\) to obtain high-dimensional representations \(x_a, x_t, x_v\), respectively. Each extractor comprises a convolutional layer followed by a bidirectional LSTM. Based on the binary missing indicator \(\beta_m\), the available modality set \(X_{\text{avail}}\) and the missing modality set \(X_{\text{miss}}\) are organized.

The generator consists of a high-dimensional diffusion model \(\mathrm{HDDM}_{\theta_\mu,\theta_\sigma}\), and a unimodal reconstructor \(\mathrm{UR}_{\theta_{ur}}\). Its objective is to generate missing modality representations that are consistent with the ground truth distribution with semantic alignment. Starting from random Gaussian noise \( \tilde{x}_m^{(n)} \sim \mathcal{N}(0, I)\), the HDDM performs a reverse diffusion process to iteratively generate the target distribution. At each step \(t\), the transition is defined as:
\begin{equation}
    p(\tilde{x}_m^{(t-1)} \mid \tilde{x}_m^{(t)}) = \mathcal{N}(\tilde{x}_m^{(t-1)}; \mu_{\theta_\mu}(\tilde{x}_m^{(t)}, t), \sigma_{\theta_\sigma}^2(\tilde{x}_m^{(t)},t)I),
\end{equation}
where \(\sigma_{\theta_\sigma}\) and \(\mu_{\theta_\mu}\) are predicted by neural networks with learnable parameters. To guide semantic correlation with available modalities \(X_{\text{avail}}\), we incorporate a conditional Vision Transformer~\cite{Peebles34} as the denoiser, enhanced with cross-attention at each denoising step. Specifically, the attention mechanism takes queries from \(\tilde{x}_m^{(t)}\) and keys/values from \(X_{\text{avail}}\), defined as \(\text{Attention}(Q, K, V) = \text{softmax}\left( \frac{QK^\top}{\sqrt{d}} \right) \cdot V\), where \( Q = W_Q^{(i)} \cdot \mathrm{IR}^{(i)}(\tilde{x}_m^{(t)}) \), \( K = W_K^{(i)} \cdot \mathrm{IR}^{(i)}(X_{\text{avail}}) \), \( V = W_V^{(i)} \cdot \mathrm{IR}^{(i)}(X_{\text{avail}}) \). \(\mathrm{IR}^{(i)}(x)\) denotes the intermediate representation at the \(i\)-th attention block, with \(W_Q^{(i)}, W_K^{(i)}, W_V^{(i)}\) as learnable projection matrices. The final reverse diffusion process is defined as: 
\begin{equation}
    \tilde{x}_m^{(t-1)} = \frac{1}{\sqrt{\gamma_t}} \left( \tilde{x}_m^{(t)} - \frac{\delta_t}{\sqrt{1-\gamma_t}} \epsilon_{\theta_\epsilon}(\tilde{x}_m^{(t)}, t,X_{\text{avail}}) \right) + \sigma_{\theta_\sigma}(\tilde{x}_m^{(t)}, t, X_{\text{avail}}) \, \epsilon,
\end{equation}
where \(\epsilon \sim \mathcal{N}(0, I)\), and \(\gamma_t\), \(\delta_t\) are the diffusion schedule parameters. After \(n\) reverse steps, we obtain the generated missing modality representation \(\tilde{x}_m^{(0)}\), which is expected to approximate the ground-truth missing modality representations \(x_m\). Each potentially missing modality (audio, text, or vision) is handled by a dedicated HDDM. Let \(\mathcal{L}_{\mathrm{HDDM}}\) denotes the loss function for the diffusion model. During training, we sample \(\tilde{x}_m^{(t)}\) and 
\(\tilde{x}_m^{(t-1)}\) via forward diffusion from \(x_m\). Following the objective derived in DDPM~\cite{DDPM21}, we approximate the objective as follows:

\begin{small}
\begin{align}
\mathcal{L}_{\mathrm{HDDM}} 
&= \sum_{m \in \{a, t, v\}} \beta_m \cdot D_{\text{KL}}\left(q(\tilde{x}_m^{(t-1)} \mid \tilde{x}_m^{(t)}, \tilde{x}_m^{(0)}) \,\|\, p(\tilde{x}_m^{(t-1)} \mid \tilde{x}_m^{(t)})\right) \notag \\
&\approx \sum_{m \in \{a, t, v\}} \beta_m \cdot \mathbb{E}_{\tilde{x}_m^{(0)},\, t,\, X_{\text{avail}}, \epsilon} \left[ 
\frac{\left\| \tilde{x}_m^{(t-1)} - \mu_{\theta_\mu}(\tilde{x}_m^{(t)}, t, X_{\text{avail}}) \right\|^2}{2\sigma^2_{\theta_\sigma}(\tilde{x}_m^{(t)}, t, X_{\text{avail}})}  
+ \frac{1}{2} \log \sigma^2_{\theta_\sigma}(\tilde{x}_m^{(t)}, t, X_{\text{avail}})
\right],
\label{LHDDM}
\end{align}
\end{small}

To refine the generated representations and reduce generation uncertainty, we design a unimodal reconstructor implemented using a residual network. This component aims to better align generated representations with the corresponding real samples, improving reconstruction fidelity. We obtain the recovered representations for the missing modalities as: \(X_{\text{rec}}=\{\beta_m\mathrm{UR}_{\theta_{ur}}(\tilde{x}_m^{(0)})\}\), with \(m \in \{a, t, v\}\). The loss is computed as:
\begin{equation}
    \mathcal{L}_{\mathrm{UR}} = \sum_{m \in \{a, t, v\}} \beta_m \| \mathrm{UR}_{\theta_{ur}}(\tilde{x}_m^{(0)}) - x_m \|^2.
    \label{LUR}
\end{equation}

\subsubsection{Multimodal Fusion Recovery}
\label{Multimodal Fusion Recovery}
The fused multimodal representations are obtained from a shared multimodal fusion network. We define the multimodal input of the ground truth as: \(X_{\text{M}_{\text{gt}}} = X_{\text{avail}} + X_{\text{miss}}\), and the recovered multimodal input as: \(X_{\text{M}_{\text{rec}}} = X_{\text{avail}} + X_{\text{rec}}\). We then obtain the corresponding fused representations: \(f_{\text{M}_{\text{gt}}}=F_{\theta_{f}}(X_{\text{M}_{\text{gt}}}), f_{\text{M}_{\text{rec}}}=F_{\theta_{f}}(X_{\text{M}_{\text{rec}}}),\)
where \(F_{\theta_{f}}(x)\) denotes the multimodal fusion network following the architecture of previous multimodal transformers~\cite{27}. 

We employ adversarial learning between the multimodal reconstructor \(\mathrm{MR}_{\theta_{mr}}\) and the discriminator \(D_{\theta_d}\). To align the recovered representation \(f_{\text{M}_{\text{rec}}}\) with the ground truth \(f_{\text{M}_{\text{gt}}}\) at the feature level, the multimodal reconstructor, which is formulated as a residual connection network, refines \(f_{\text{M}_{\text{rec}}}\) as \(\tilde{f}_{\text{M}_{\text{rec}}}=\mathrm{MR}_{\theta_{mr}}(f_{\text{M}_{\text{rec}}})\). The reconstruction loss is given by: 
\begin{equation}
    \mathcal{L}_{{rec}}= \sum_{i=1}^{N} \| \tilde{f}^{(i)}_{\text{M}_{\text{rec}}} - f^{(i)}_{\text{M}_{\text{gt}}} \|^2,
    \label{Lrec}
\end{equation}
where \(i\) indexes the training instances and \(N\) denotes the total number of instances. Although deterministic reconstruction losses reduce feature-level discrepancies, they are insufficient to model the complex distribution of high-dimensional multimodal representations, unable to compensate for semantic degradation caused by missing modalities. Thus, we design a discriminator implemented as a multi-layer perceptron, which distinguishes between recovered and ground-truth representation, producing outputs \(I^{(i)}_{\text{rec}}=D_{\theta_d}(\tilde{f}_{\text{M}_{\text{rec}}})\), \(I^{(i)}_{\text{gt}}=D_{\theta_d}(f_{\text{M}_{\text{gt}}})\). The discriminator is trained to classify \(f^{(i)}_{\text{gt}}\) as real and \(\tilde{f}^{(i)}_{\text{rec}}\) as fake, with the following objective:
\begin{equation}
    \mathcal{L}_D = -\frac{1}{N} \sum_{i=1}^{N} \log I^{(i)}_{\text{gt}} - \frac{1}{N} \sum_{i=1}^{N} \log \left(1 - I^{(i)}_{\text{rec}} \right).
    \label{LD}
\end{equation}
The multimodal reconstructor is trained to make \(\tilde{f}_{\text{M}_{\text{rec}}}\) indistinguishable from \({f}_{\text{M}_{\text{gt}}}\) by the discriminator, with the following objective:
\begin{equation}
    \mathcal{L}_{{adv}}=-\frac{1}{N} \sum_{i=1}^{N} \log I^{(i)}_{\text{rec}}.
    \label{Ladv}
\end{equation}
The total MR loss combines reconstruction (Equation~\ref{Lrec}) and adversarial (Equation~\ref{Ladv}) terms:
\begin{equation}
    \mathcal{L}_{\mathrm{MR}}=\lambda_{al}\mathcal{L}_{{adv}}+(1-\lambda_{al})\mathcal{L}_{{rec}},
    \label{LMR}
\end{equation}
where \(\lambda_{\text{adv}} \in [0,1]\) balances semantic alignment and feature reconstruction. Importantly, the adversarial learning scheme also promotes the fusion network to produce more robust and discriminative multimodal representations.

The classifier \(C_{\theta_c}\) is trained to perform emotion recognition based on the fused multimodal representation, which is implemented as a feed-forward network. During training, it receives both the reconstructed representation \(\tilde{f}^{(i)}_{\text{rec}}\) and the ground-truth representation \(f^{(i)}_{\text{gt}}\) as inputs to enhance robustness and generalization. The corresponding predictions are denoted as \(y^{(i)}_{\text{rec}} = C_{\theta_c}(\tilde{f}^{(i)}_{\text{rec}}), \quad y^{(i)}_{\text{gt}} = C_{\theta_c}(f^{(i)}_{\text{gt}}),\)
The classifier losses are defined as:
\begin{equation}
    \mathcal{L}_{{C}_1}=\sum_{i=1}^{N}\|y^{(i)} -y^{(i)}_{\text{rec}}\|^2, \quad \mathcal{L}_{{C}_2}=\sum_{i=1}^{N}\|y^{(i)} -y^{(i)}_{\text{gt}}\|^2,
    \label{LC}
\end{equation}
where \({(i)}\) denotes the \(i\)-th instance, \(N\) is the total instance counts, \(\lambda_{c}\) is a hyperparameter balancing the contribution of each term.

\subsection{Multi-Stage Optimization Strategy}
\label{Multi-Stage Optimization Strategy}
We adopt a multi-stage optimization strategy to enhance the training stability and efficiency, as shown in Figure~\ref{fig 1}(c). This strategy, in which the losses of all modules are decomposed into three distinct optimization objectives, fed the input into the model three times during each training epoch with distinct objectives. The stage 1 is designed to enhance the model’s capability to denoise and recover representation for missing modalities based on the generator loss (Equation~\ref{LHDDM} and~\ref{LUR}). The objective is formulated as follows:
\begin{equation}
    \mathcal{L}_{S1}=\frac{\mathcal{L}_{\mathrm{HDDM}} + \lambda_g \mathcal{L}_{\mathrm{UR}}}{1 + \lambda_g}.
    \label{stage 1}
\end{equation}
The stage 2 focuses on improving the fusion of multimodal representation based on the adversarial learning loss (Equation~\ref{LD} and~\ref{LMR}):
\begin{equation}
    \mathcal{L}_{S2}=\mathcal{L}_{D} +\lambda_{al}\mathcal{L}_{{adv}}+(1-\lambda_{al})\mathcal{L}_{{rec}}.
    \label{stage 2}
\end{equation}
The stage 3 aims to strengthen the model’s robustness in performing emotion classification based on the classifier loss (Equation~\ref{LC}):
\begin{equation}
    \mathcal{L}_{S3}=\lambda_c\mathcal{L}_{{C}_1}+(1-\lambda_c)\mathcal{L}_{{C}_2},
    \label{stage 3}
\end{equation}
where \(\lambda_g\), \(\lambda_{al}\), and \(\lambda_c\) control the weights of the generative, adversarial, and classifier losses, respectively. Importantly, in this strategy, the loss at each stage is used to jointly optimize all involved modules without freezing any parameters. \textbf{Algorithm 1} illustrates the detailed procedure for implementing the proposed optimization strategy.

\begin{algorithm}

\caption{Multi-Stage Optimization Strategy}
\begin{algorithmic}[1]
\State Initialize model parameters $\theta_{fe},\theta_\mu,\theta_\sigma,\theta_{ur},\theta_f,\theta_d,\theta_{mr},\theta_c$.
\State Define optimizer $\mathcal{O}_{d}$ for discriminator, optimizer $\mathcal{O}$ for the remaining modules.

\For{each training epoch}
    \State \textbf{Stage 1:} Optimization of $\theta_{fe},\theta_\mu,\theta_\sigma,\theta_{ur}$
    \State \hspace{1em} Pass \( \tilde{x}_m^{(n)} \) through HDDM and UR with \(X_{\textbf{avail}}\).
    \State \hspace{1em} Compute $\mathcal{L}_{S1}$(Equation~\ref{stage 1}) and update target $\theta$ using $\mathcal{O}$.
    
    \State \textbf{Stage 2:} Optimization of $\theta_{fe},\theta_\mu,\theta_\sigma,\theta_{ur},\theta_f,\theta_d,\theta_{mr}$
    \State \hspace{1em} Repeat step 5. Pass \(X_{\text{M}_{\text{rec}}}\) and \(X_{\text{M}_{\text{gt}}}\) through Fusion, MR and Discriminator.
    \State \hspace{1em} Compute $\mathcal{L}_{D}$(Equation~\ref{LD}) and Update $\theta_d$ using $\mathcal{O}_d$.
    \State \hspace{1em} Pass \(\tilde{f}_{\text{M}_{\text{rec}}}\) and \(f_{\text{M}_{\text{gt}}}\) through Discriminator.
    \State \hspace{1em} Compute $\mathcal{L}_{MR}$(Equation~\ref{LMR}) and update  $\theta_{fe},\theta_\mu,\theta_\sigma,\theta_{ur},\theta_f,\theta_{mr}$ using $\mathcal{O}$.
    
    \State \textbf{Stage 3:} Optimization of  $\theta_{fe},\theta_\mu,\theta_\sigma,\theta_{ur},\theta_f,\theta_{mr},\theta_c$
    \State \hspace{1em} Repeat step 8. Pass \(\tilde{f}_{\text{M}_{\text{rec}}}\) and \(f_{\text{M}_{\text{gt}}}\) through Emotion Classifier.
    \State \hspace{1em} Compute $\mathcal{L}_{S3}$(Equation~\ref{stage 3}) and update target $\theta$ using $\mathcal{O}$.
\EndFor

\end{algorithmic}
\end{algorithm}

\section{Experiments}
\subsection{Datasets and Metrics}
\label{Datasets and Metrics}
\textbf{Dataset.} We evaluate our method on two standard MSA benchmarks. \textbf{CMU-MOSI}~\cite{MOSI28} includes 2,199 video segments from 93 YouTube movie reviews, and \textbf{CMU-MOSEI}~\cite{MOSEI29} contains 22,856 annotated clips. For both, audio and vision representation are extracted using COVAREP~\cite{COVAREP30} and Facet~\cite{facet31}, while text is encoded via a pre-trained BERT~\cite{BERT32}. Sentiment labels range from -3 to 3, spanning seven levels from highly negative to highly positive.

\textbf{Metrics and Baselines.} We report binary accuracy (\(\text{ACC}_2\)), 7-class accuracy (\(\text{ACC}_7\)), and F1-score. \(\text{ACC}_2\) and \(\text{ACC}_7\) measure the proportion of correct predictions under 2 and 7 sentiment classes, respectively. F1 is the harmonic mean of Precision and Recall. We compare our proposed model with existing state-of-the-art methods. For the MOSI dataset, we adopt MTMSA~\cite{MTMSA12} and SMCMSA~\cite{SMCMSA16} as baselines for representation learning-based methods, and IMDer~\cite{IMDer13} and DiCMoR~\cite{DiCMoR17} as baselines for generative model-based methods. For the MOSEI dataset, we select MMIN~\cite{MMIN14} and GCNet~\cite{GCNet10} as baselines for representation learning-based methods, IMDer and DiCMoR as generative model-based baselines. A brief description of all baseline models is provided in Related Work of the technical appendices and supple mentary material. 

\subsection{Comparison to State-of-the-art}
\label{Comparison to State-of-the-art}
\subsubsection{Quantitative Results}
\label{Quantitative results}
Following~\cite{DiCMoR17,GCNet10,MMIN14,IMDer13}, we evaluate our method under two common IMER settings: random modality missing and modality-specific availability. For random modality missing, available modalities are randomly selected in each sample. The missing rate (MR) is defined as \(\text{MR} = 1 - \frac{1}{N \times M} \sum_{i=1}^{N} m^{(i)}\), where \(N\) is the number of samples, \(M\) the total modalities, and \(m^{(i)}\) the available modalities in the \(i\)-th sample. MR ranges from 0.0 to 0.7, ensuring at least one modality per sample. For modality-specific availability, we simulate missing patterns where only one (\{a\}, \{t\}, \{v\}) or two modalities (\{a, t\}, \{a, v\}, \{t, v\}) are available. For the MOSI and MOSEI datasets, we set \(\lambda_{g}=1\) and \(\lambda_{al}=0.5\); the value of \(\lambda_{c}\) is set to 0.4 and 0.6, respectively.

\begin{table}[ht]
  \caption{Results under random modality missing. The values reported in each cell denote \(\text{ACC}_2\)/F1. \textbf{Bold} is the best.}
  \label{random miss}
  \centering
  \resizebox{\textwidth}{!}{
    \begin{tabular}{c|c|cccccccc}
      \toprule
      \multirow{2}{*}{\textbf{Dataset}} & \multirow{2}{*}{\textbf{Models}} & \multicolumn{8}{c}{\textbf{Missing Rates}} \\
      \cmidrule{3-10}
      & & 0 & 0.1 & 0.2 & 0.3 & 0.4 & 0.5 & 0.6 & 0.7  \\
      \midrule
      \multirow{5}{*}{MOSI}
      & IMDer~\cite{IMDer13}   & 85.70/85.60 & 84.90/84.80 & 83.50/83.40 & 81.20/81.00 & 78.60/78.50 & 76.20/75.90 & 74.70/74.00 & 71.90/71.20  \\
      & DiCMoR~\cite{DiCMoR17}  & 85.70/85.60 & 83.90/83.90 & 82.10/82.00 & 80.40/80.20 & 77.90/77.70 & 76.70/76.40 & 73.30/73.00 & 71.10/70.80  \\
      & MTMSA~\cite{MTMSA12}   & 84.89/58.12 & 84.37/57.43 & 81.25/55.71 & 78.12/52.82 & 76.04/51.32 & 73.43/51.72 & - & -  \\
      & SMCMSA~\cite{SMCMSA16}  & 85.44/59.32 & 84.89/58.34 & 83.60/57.07 & 80.01/55.35 & 77.88/53.76 & 75.38/52.71 & 73.23/52.07 & 70.14/50.46  \\
      & RoHyDR (Ours)    & \textbf{86.89/86.80} & \textbf{86.57/86.50} & \textbf{85.32/84.70} & \textbf{83.23/82.97} & \textbf{81.76/81.03} & \textbf{80.23/79.67} & \textbf{78.94/77.83} & \textbf{76.63/75.46}  \\
      \midrule
       \multirow{5}{*}{MOSEI}
      & MMIN~\cite{MMIN14}   & 84.71/84.11 & 82.09/81.42 & 80.02/79.30 & 77.96/75.47 & 74.60/72.21 & 72.20/70.55 & 70.90/69.08 & 70.08/69.43  \\
      & GCNet~\cite{GCNet10}  & 85.04/84.63 & 82.90/82.03 & 80.17/78.93 & 78.04/76.23 & 75.92/74.16 & 73.76/71.97 & 72.30/70.54 & 71.84/69.90  \\
      & IMDer~\cite{IMDer13}   & 85.34/85.10 & 84.76/\textbf{83.53} & 82.54/80.77 & 80.95/78.19 & 78.33/75.62 & 77.25/74.42 & 75.84/73.94 & 75.00/72.97  \\
      & DiCMoR~\cite{DiCMoR17}   & 85.34/85.10 & 83.63/82.81 & 81.26/79.88 & 79.75/77.38 & 77.63/75.02 & 75.80/73.72 & 74.51/72.87 & 73.07/71.66  \\
      & RoHyDR (Ours)    & \textbf{86.10/85.43} & \textbf{84.96}/83.04 & \textbf{83.07/81.70} & \textbf{81.87/80.07} & \textbf{80.76/77.03} & \textbf{79.97/76.87} & \textbf{78.03/74.83} & \textbf{76.89/73.42}  \\
      \bottomrule
    \end{tabular}
  }
\end{table}

\begin{table}[ht]
\vspace{-1em}
  \caption{Results under modality-specific availability. The values reported in each cell denote \(\text{ACC}_2\)/F1. \textbf{Bold} is the best. }
  \label{available conditions}
  \centering
  \resizebox{\textwidth}{!}{
    \begin{tabular}{c|c|ccccccc}
      \toprule
      \multirow{2}{*}{\textbf{Dataset}} & \multirow{2}{*}{\textbf{Models}} & \multicolumn{7}{c}{\textbf{Available Conditions}} \\
      \cmidrule{3-9}
      & & \{\(a\)\} & \{\(t\)\} & \{\(v\)\} & \{\(a, t\)\} & \{\(a, v\)\} & \{\(t, v\)\} & \{\(a, t, v\)\}  \\
      \midrule
      \multirow{5}{*}{MOSI}
      & IMDer~\cite{IMDer13}   & 62.00/62.20 & 84.80/84.70 & 61.30/60.80 & 85.40/85.30 & 63.60/63.40 & 85.50/85.40 & 85.70/85.60 \\
      & DiCMoR~\cite{DiCMoR17}  & 60.50/60.80 & 84.50/84.40 & 62.20/60.20 & 85.50/85.50 & 64.00/\textbf{63.50} & 85.50/85.40 & 85.70/85.60 \\
      & MTMSA~\cite{MTMSA12}   & 60.93/40.50 & 78.64/53.46 & 57.29/35.02 & 82.29/55.54 & 61.97/41.12 & 80.72/54.73 & 84.89/58.12 \\
      & SMCMSA~\cite{SMCMSA16}  & 56.25/35.01 & 84.38/57.12 & 55.21/31.81 & 83.85/56.34 & 58.33/38.59 & 83.33/56.34 & 85.94/59.74 \\
      & RoHyDR (Ours) & \textbf{63.89/62.73} & \textbf{86.11/84.88} & \textbf{63.07/62.31} & \textbf{86.54/86.37} & \textbf{64.59}/63.02 & \textbf{86.42/86.04} & \textbf{86.89/86.80} \\
      \midrule
       \multirow{5}{*}{MOSEI}
      & MMIN~\cite{MMIN14}   & 56.32/55.94 & 83.67/83.42 & 56.73/56.44 & 83.74/83.44 & 62.67/61.32 & 83.65/83.51 & 70.90/69.08  \\
      & GCNet~\cite{GCNet10}  & 59.73/59.24 & 82.87/82.43 & 60.82/60.64 & 84.23/84.04 & 64.23/63.77 & 84.17/83.92 & 84.83/84.57  \\
      & IMDer~\cite{IMDer13}   & 63.80/60.60 & 84.50/84.50 & 63.90/63.60 & 85.10/85.10 & 64.90/63.50 & 85.00/85.00 & 85.10/85.10 \\
      & DiCMoR~\cite{DiCMoR17}   & 62.90/60.40 & 84.20/84.30 & 63.60/63.60 & 85.00/84.90 & 65.20/64.40 & 84.90/84.90 & 85.10/85.10 \\
      & RoHyDR (Ours)    & \textbf{65.03/63.79} & \textbf{84.82/84.63} & \textbf{64.23/63.43} & \textbf{85.43/85.22} & \textbf{66.32/65.07} & \textbf{85.62/85.14} & \textbf{86.10/85.43}  \\
      \bottomrule
    \end{tabular}
  }
  % \vspace{-1em}
\end{table}
All experiments are repeated five times, and we report test results in Tables~\ref {random miss} and~\ref{available conditions}. We have the following observations:
\begin{enumerate}[(1)]
    \item RoHyDR consistently outperforms all baselines across both MER datasets and IMER settings, demonstrating strong generalization. Compared to representation learning-based methods~\cite{GCNet10,MTMSA12,SMCMSA16,MMIN14}, it achieves superior performance by enabling fine-grained recovery of missing modalities and effectively conditioning on available inputs. Its advantage over generative model-based methods~\cite{IMDer13,DiCMoR17} further indicates its capacity for recovery at the feature level and semantic level in multimodal fusion.
    
    \item Under the random modality missing setting, RoHyDR demonstrates strong robustness to high missing rates. As shown in Table~\ref{random miss}, on MOSI, its accuracy gain over baselines is modest 1.29\%--2.00\% at a 0.0 missing rate, but increases to 4.24\%--5.71\% at 0.6 and 4.73\%--6.49\% at 0.7. A similar trend holds for MOSEI, with gains reaching up to 6.81\% at 0.7. Moreover, as the missing rate on MOSI rises from 0 to 0.7, baseline accuracy drops by 13.80\%--15.30\%, while RoHyDR’s decline is only 10.36\%, yielding a relative improvement of 24.93\%--32.29\%.

    \item Under the modality-specific availability setting, RoHyDR effectively recovers both vision and audio information, as shown in Table~\ref{available conditions}. When only the audio \{t, v\} or vision \{a, t\} modality is available, it achieves the highest accuracy and F1 scores across all baselines. Moreover, even when two modalities are missing, RoHyDR maintains superior performance, with accuracy improvements ranging from 0.62\% to 7.47\% on both MER datasets. These results highlight the strong resilience of the model under highly incomplete multimodal conditions.
\end{enumerate}

\subsubsection{Qualitative Results}
\label{Qualitative results}

\begin{figure*}[!ht]
% \vspace{-2.0em}
\centering
\subfloat[\scriptsize ]{
    \includegraphics[width=0.24\linewidth]{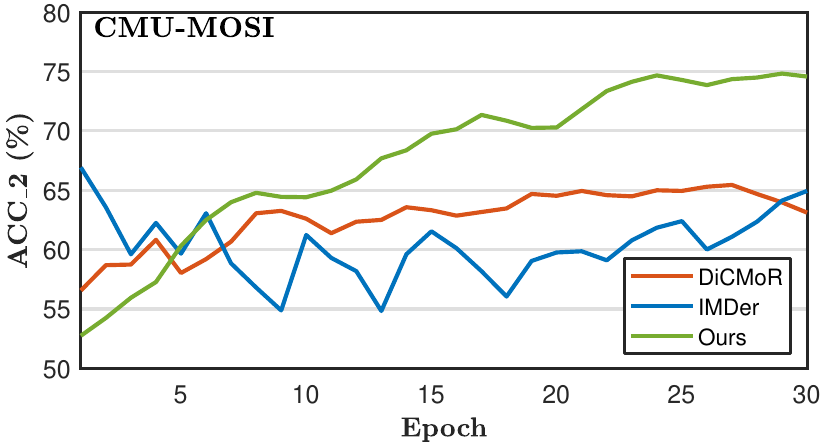}
    \label{mositest}
}
\subfloat[\scriptsize ]{
    \includegraphics[width=0.24\linewidth]{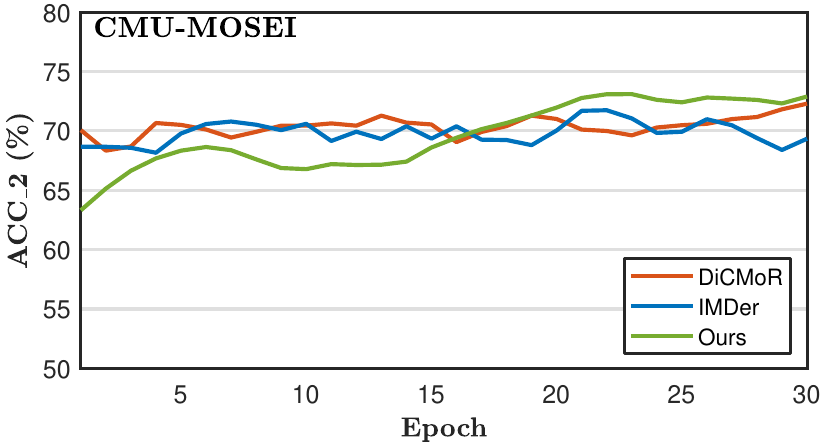}
    \label{moseitest}
}
\subfloat[\scriptsize ]{
    \includegraphics[width=0.24\linewidth]{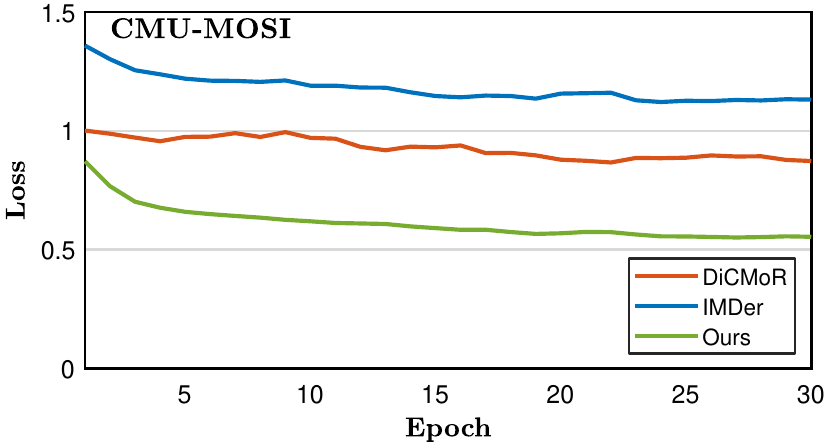}
    \label{mosiloss}
}
\subfloat[\scriptsize ]{
    \includegraphics[width=0.24\linewidth]{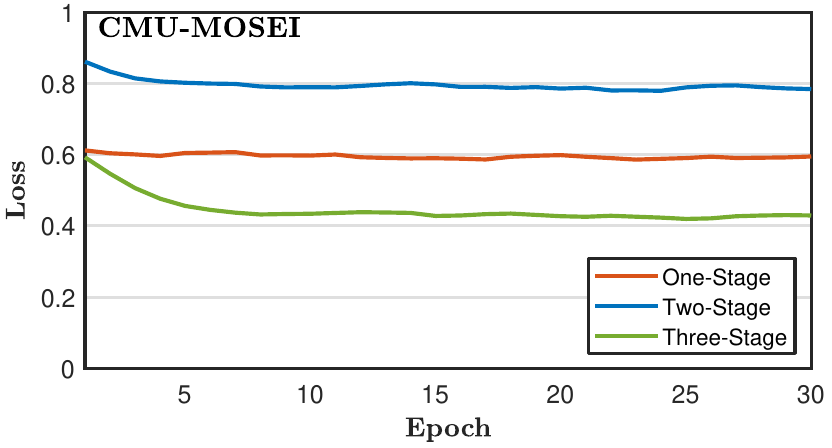}
    \label{moseiloss}
}\\
\subfloat[\scriptsize ]{
    \includegraphics[width=0.24\linewidth]{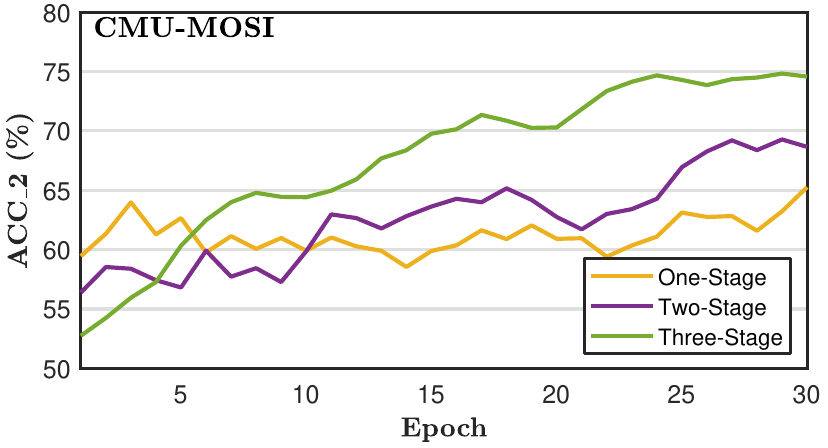}
    \label{mosistrategytest}
}
\subfloat[\scriptsize ]{
    \includegraphics[width=0.24\linewidth]{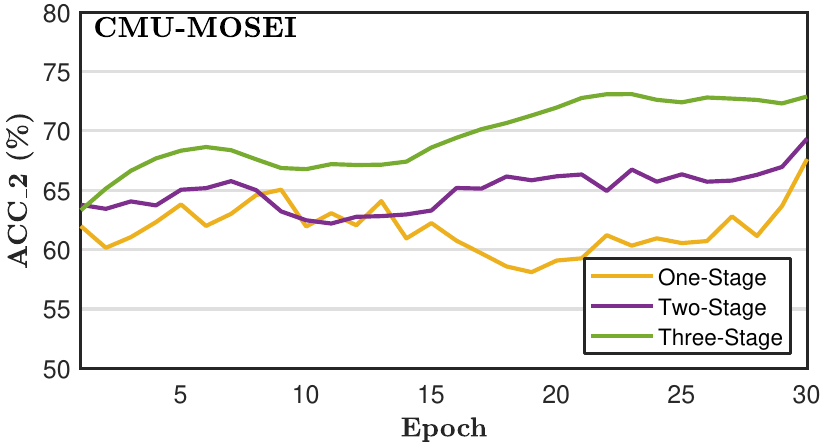}
    \label{moseistrategytest}
}
\subfloat[\scriptsize ]{
    \includegraphics[width=0.24\linewidth]{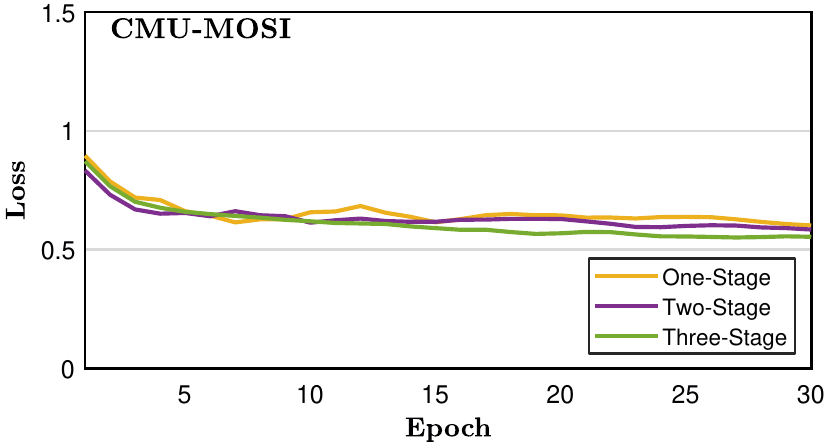}
    \label{mosistrategyloss}
}
\subfloat[\scriptsize ]{
    \includegraphics[width=0.24\linewidth]{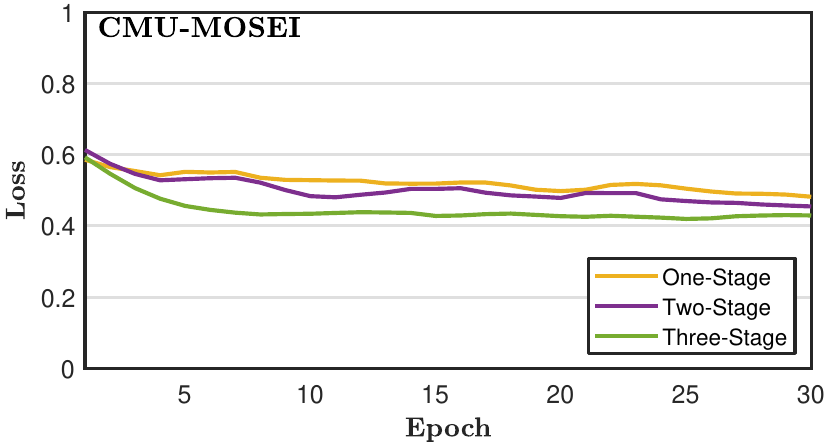}
    \label{moseistrategyloss}
}
\caption{Visualization of training stability and efficiency. (a–d) illustrate the comparison between RoHyDR and state-of-the-art methods. (e–h) demonstrate the ablation study of optimization strategy. }
\label{fig 3}
% \vspace{-1.5em}
\end{figure*}

We assess training stability and efficiency on MOSI and MOSEI with a missing rate of 0.7. RoHyDR is compared against generative modal-based baselines IMDer~\cite{IMDer13} and DiCMoR~\cite{DiCMoR17} using identical training hyperparameters (e.g., learning rate), while preserving their official configurations otherwise. All models are trained for 30 epochs. For stability, we track \(\text{ACC}_2\) over epochs. As shown in Figure~\ref{fig 3}(a-b), RoHyDR achieves more consistent accuracy improvements during training, with notably fewer fluctuations. For efficiency, Figure~\ref{fig 3}(c-d) illustrates faster and smoother convergence of classifier loss, reflecting more effective optimization dynamics.

\subsection{Ablation Study}
\subsubsection{Quantitative Analysis}
We evaluate the core components involved in unimodal representation recovery and multimodal fusion recovery,  including the high-dimensional diffusion model  \(\mathrm{HDDM}_{\theta_\mu,\theta_\sigma}\), unimodal reconstructor \(\mathrm{UR}_{\theta_{ur}}\), discriminator \(D_{\theta_d}\) and multimodal reconstructor \(\mathrm{MR}_{\theta_{mr}}\). The results are shown in Table~\ref{ab Feature recovery} and~\ref{ab Semantic recovery}. Key findings are summarized as follows:  (1) \(\mathrm{HDDM}_{\theta_\mu,\theta_\sigma}\) and \(\mathrm{UR}_{\theta_{ur}}\) are critical to unimodal recovery. The former uses available modalities as conditional inputs to generate semantically aligned and distribution-consistent representations, while the latter improves fidelity and mitigates generative instability. (2) \(\mathrm{MR}_{\theta_{mr}}\) and \(D_{\theta_d}\) are essential for multimodal fusion recovery. \(\mathrm{MR}_{\theta_{mr}}\) explicitly reconstructs fused features, and \(D_{\theta_d}\) provides adversarial guidance for semantic alignment. This dual mechanism enhances robustness by reducing susceptibility to input perturbations.

\begin{table}[!ht]
% \vspace{-1.0em}
\centering
\begin{minipage}{0.5\textwidth}
  \centering
  \caption{Ablation study of unimodal representation recovery module under random modality missing. Each cell reports the average performance over missing rate values from 0.0 to 0.7.}
  \label{ab Feature recovery}
  \resizebox{\textwidth}{!}{
    \begin{tabular}{c|cc|ccc}
      \toprule
      Dataset& \(\mathrm{HDDM}_{\theta_\mu,\theta_\sigma}\) & \(\mathrm{UR}_{\theta_{ur}}\) &\(\text{ACC}_2\) & \(\text{ACC}_7\) & \(\text{F1}\)    \\
      \midrule
      \multirow{4}{*}{MOSI}
      &\(\times\) &\(\times\)   & 77.83 & 37.02 & 77.21\\
      &\(\times\) &\(\checkmark\)   & 78.93 & 38.62 & 78.11\\
      &\(\checkmark\) &\(\times\)   & 80.51 & 40.24 & 79.78\\ 
      &\(\checkmark\) &\(\checkmark\)   & \textbf{82.45} & \textbf{42.03} & \textbf{81.87} \\
      \bottomrule
    \end{tabular}
  }
\end{minipage}
\hfill
\begin{minipage}{0.48\textwidth}
  \centering
  \caption{Ablation study of multimodal fusion recovery under random modality missing. Each cell reports the average performance over MR values from 0.0 to 0.7.}
  \label{ab Semantic recovery}
  \resizebox{\textwidth}{!}{
    \begin{tabular}{c|cc|ccc}
      \toprule
      Dataset& \(D_{\theta_d}\) & \(\mathrm{MR}_{\theta_{mr}}\) &\(\text{ACC}_2\) & \(\text{ACC}_7\) & \(\text{F1}\)    \\
      \midrule
      \multirow{4}{*}{MOSI}
      &\(\times\) &\(\times\)   & 80.01 & 40.76 & 79.52\\
      &\(\times\) &\(\checkmark\)   & 80.94 & 41.02 & 80.24\\
      &\(\checkmark\) &\(\times\)   & 81.74 & 41.49 & 81.15\\ 
      &\(\checkmark\) &\(\checkmark\)   & \textbf{82.45} & \textbf{42.03} & \textbf{81.87} \\
      \bottomrule
    \end{tabular}
  }
\end{minipage}
% \vspace{-1.5em}
\end{table}

We also evaluate performance under different optimization strategies. One-stage optimization strategy corresponds to the conventional approach, formulated as \(\mathcal{L}_{S1}^{(1)} = \lambda_g (\mathcal{L}_\mathrm{HDDM} + \mathcal{L}_\mathrm{UR}) +  (\mathcal{L}_{D} + \mathcal{L}_\mathrm{MR}) + \lambda_c\mathcal{L}_{{C}_1}+(1-\lambda_c)\mathcal{L}_{{C}_2}\), where \(\lambda_g\), \(\lambda_{al}\) and \(\lambda_c\) control the weights of the generative, adversarial and classifier losses, respectively.
The two-stage optimization strategy decomposes the process into two functional phases: the first focuses on enhancing the model's ability to recover missing representation, while the second is designed to improve joint representation learning and robust classification performance, iteratively optimize  \(\mathcal{L}_{S1}^{(2)}=\frac{\mathcal{L}_\mathrm{HDDM} + \lambda_g \mathcal{L}_\mathrm{UR}}{1 + \lambda_g}\) and  \(\mathcal{L}_{S2}^{(2)}=(\mathcal{L}_{D} + \mathcal{L}_\mathrm{MR}) + \lambda_c\mathcal{L}_{{C}_1}+(1-\lambda_c)\mathcal{L}_{{C}_2}\) in sequence. We set the default values to \(\lambda_{g}=1\), \(\lambda_{al}=0.5\), and \(\lambda_{c}=0.5\). Results are shown in the table~\ref{ab Training Strategy}. In conclusion, the proposed multi-stage optimization strategy plays a pivotal role in enhancing RoHyDR’s performance. It formulates the IMER task as a sequence of three temporally ordered sub-objectives and progressively optimizes them within each training epoch. Unlike the conventional one-stage strategy that treats emotion recognition as a unified objective, our approach explicitly disentangles the subtasks and models their temporal dependencies. This decomposition reduces gradient interference across tasks and stabilizes multi-task training, leading to substantial performance gains.

\begin{table}[!ht]
% \vspace{-1.0em}
  \caption{Ablation study of multi-stage optimization strategy under random modality missing. Each cell reports the average performance over missing rate values from 0.0 to 0.7. \textbf{Bold} is the best.}
  \label{ab Training Strategy}
  \centering
  
  \resizebox{0.5\textwidth}{!}{
    \begin{tabular}{c|c|ccc}
      \toprule
      Dataset& Training Strategy &\(\text{ACC}_2\) & \(\text{ACC}_7\) & \(\text{F1}\)    \\
      \midrule
      \multirow{3}{*}{MOSI}
      &One-Stage  & 81.07 & 40.79 & 80.67\\
      &Two-Stage   & 82.08 & 41.87 & 81.55\\
      &Ours   & \textbf{82.45} & \textbf{42.03} & \textbf{81.87} \\ 
      \bottomrule
    \end{tabular}
  }
  % \vspace{-1.5em}
\end{table}

\subsubsection{Qualitative Analysis}
\begin{figure*}[!ht]
% \vspace{-2.0em}
\centering
\subfloat[\scriptsize \(\lambda_g\) ]{
    \includegraphics[width=0.32\linewidth]{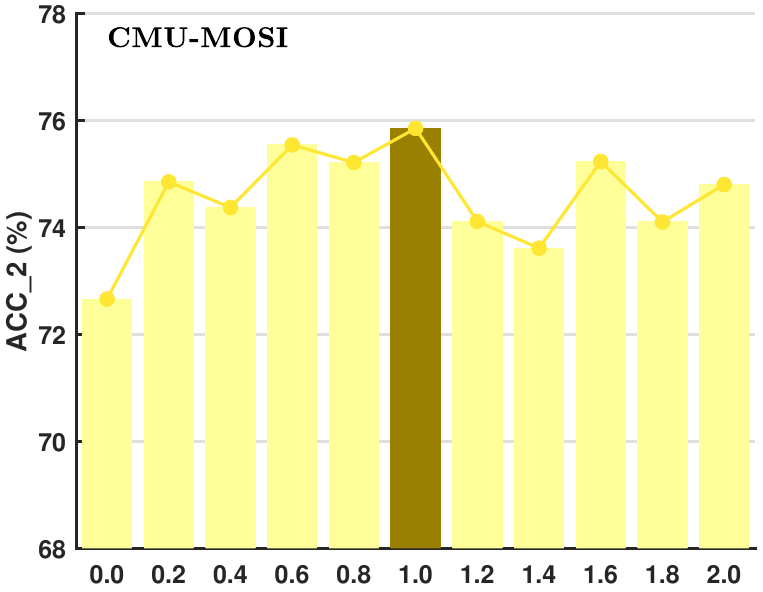}
    \label{generativemosi}
}
\subfloat[\scriptsize \(\lambda_{al}\)]{
    \includegraphics[width=0.32\linewidth]{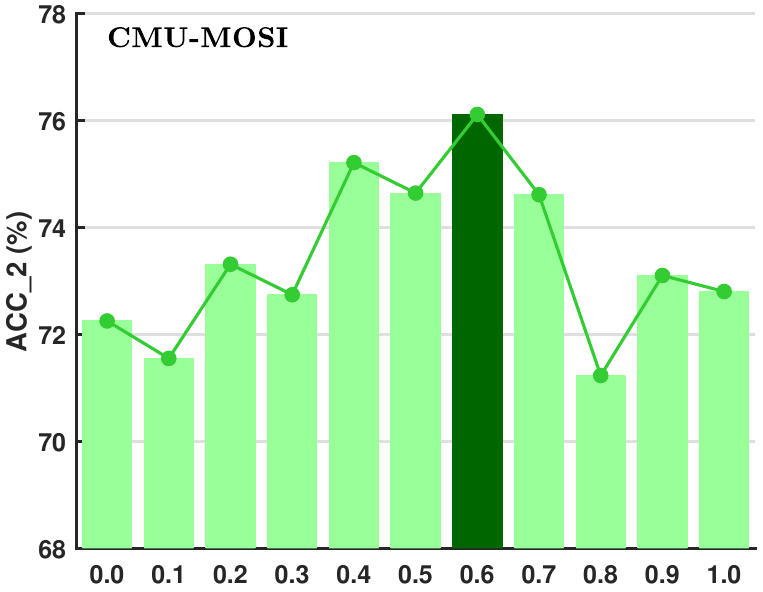}
    \label{adversarialmosi}
}
\subfloat[\scriptsize \(\lambda_{c}\)]{
    \includegraphics[width=0.32\linewidth]{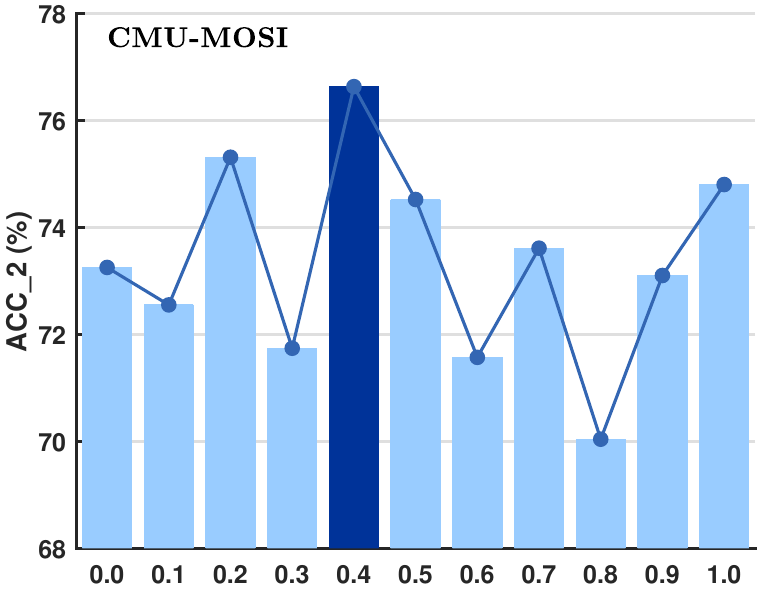}
    \label{classifiermosi}
}
\\
\subfloat[\scriptsize \(\lambda_g\)]{
    \includegraphics[width=0.32\linewidth]{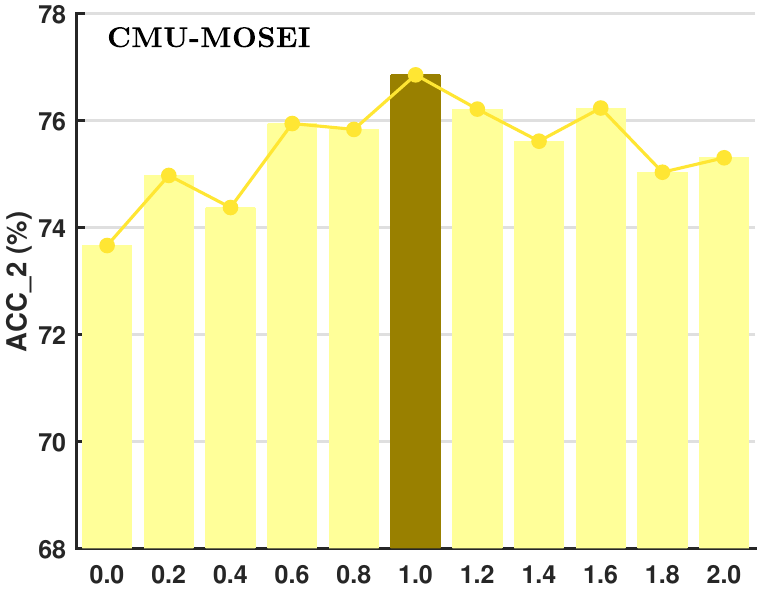}
    \label{generativemosei}
}
\subfloat[\scriptsize \(\lambda_{al}\)]{
    \includegraphics[width=0.32\linewidth]{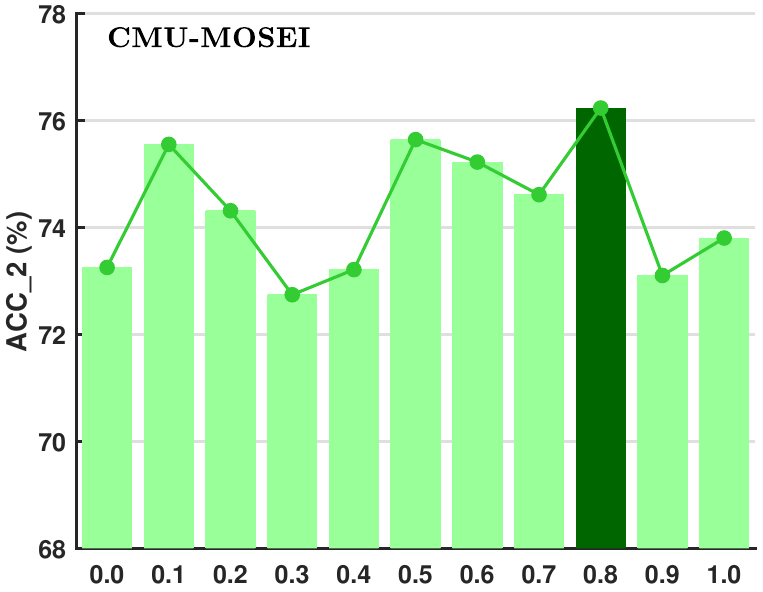}
    \label{adversarialmosei}
}
\subfloat[\scriptsize \(\lambda_{c}\)]{
    \includegraphics[width=0.32\linewidth]{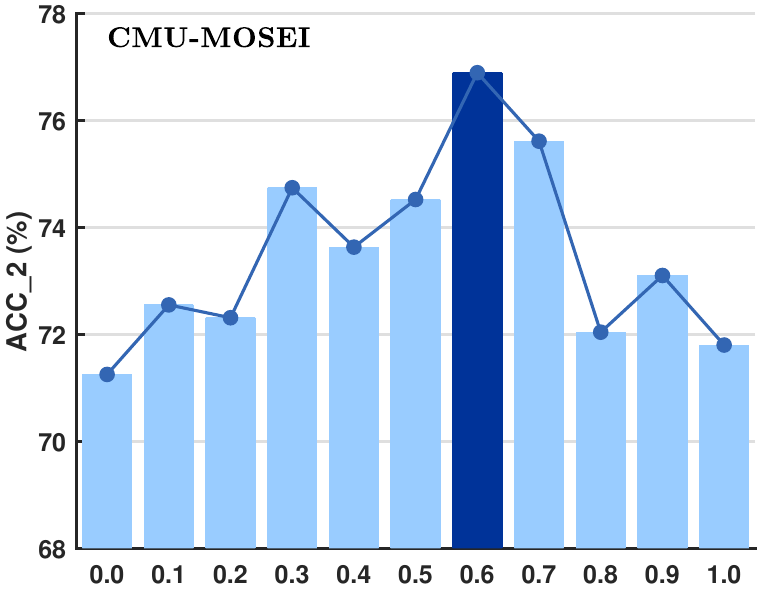}
    \label{classifiermosei}
}
\caption{ Hyper-parameter analysis. (a) and (d) analyze the effect of \(\lambda_{g}\), (b) and (e) analyze \(\lambda_{al}\), (c) and (f) examine the impact of \(\lambda_{c}\).}
\label{hyper-parameter analysis}
 % \vspace{-1.5em}
\end{figure*}
We visualize the training stability and efficiency of each optimization strategy with the same settings in Section~\ref{Qualitative results}, as shown in Figure~\ref{fig 3}(e-h), where Three-Stage denotes our proposed strategy. We further conduct hyper-parameter analysis on the generative loss, adversarial loss and classification loss, focusing on \(\lambda_{g}\), \(\lambda_{al}\) and \(\lambda_{c}\)(Equation~\ref{stage 1}-\ref{stage 3}). We set the default values to \(\lambda_{g}=1\), \(\lambda_{al}=0.5\), and \(\lambda_{c}=0.5\). Our analysis presented in Figure~\ref{hyper-parameter analysis} demonstrates the following: (1) Our model yields optimal performance on both MER datasets when \(\lambda_{g}=1\), suggesting that \(\mathcal{L}_\mathrm{HDDM}\) and \(\mathcal{L}_\mathrm{UR}\) play equally critical roles in unimodal representation recovery. When \(\lambda_{g} < 1\), insufficient emphasis on \(\mathcal{L}_\mathrm{UR}\) leads to less realistic reconstructions and weaker robustness. Conversely, setting \(\lambda_{g} > 1\) overemphasizes \(\mathcal{L}_\mathrm{UR}\), amplifying instability from generative outputs and undermining semantic alignment guided by \(\mathcal{L}_\mathrm{HDDM}\). This highlights the necessity of a balanced optimization between generative consistency and reconstruction fidelity in Stage 1.
(2) On the MOSI dataset, the best performance is obtained at \(\lambda_{al}=0.6\), while on MOSEI it occurs at \(\lambda_{al}=0.8\), suggesting that \(\mathcal{L}_{adv}\) plays a more critical role in multimodal fusion recovery. 
(3) The optimal \(\lambda_{c}\) values are 0.4 and 0.6 for MOSI and MOSEI, respectively. Although this does not isolate the contributions of \(\mathcal{L}_{C_1}\) and \(\mathcal{L}_{C_2}\), a comparison with \(\lambda_{c}=1\) shows that using the classification output of the complete multimodal fused representation as a supervisory signal effectively enhances model performance.
(4)The differences in optimal \(\lambda_{al}\) and \(\lambda_{c}\) reflect dataset complexity. MOSEI’s diverse, fine-grained annotations demand stronger semantic alignment and supervision, leading to higher values. In contrast, MOSI’s smaller scale allows effective learning with weaker adversarial and classification signals.

\section{Conclusions}
In this paper, we address the challenges of incomplete multimodal emotion recognition by proposing a novel method, called robust hybrid diffusion recovery (RoHyDR).
Specifically, RoHyDR employs a high-dimensional diffusion model to generate semantically aligned and distribution-consistent representation of the missing modalities and a unimodal reconstructor to alleviate uncertainty arising from the generative process.
Secondly, to facilitate the recovery of emotionally rich fused representation from incomplete multimodal inputs, we employ adversarial learning, where a discriminator guides the fusion process toward semantic alignment with complete multimodal representation, enhancing the ability of both the fusion network and the multimodal reconstructor to generate robust multimodal representation.
Finally, a multi-stage optimization strategy is adopted to ensure training stability and efficiency, improving performance. Extensive quantitative and qualitative experiments consistently validate the effectiveness of our proposed method.
However, despite its improved performance through the multi-stage optimization strategy, our method still faces certain limitations. The strategy introduces increased time complexity, resulting in longer training times per epoch. In future work, we aim to enhance the time efficiency of the optimization process.

% \bibliographystyle{plain}  % 或者使用 unsrt, IEEEtran 等样式
% % \bibliographystyle{plainnat} 
% \bibliography{References} 

\begin{thebibliography}{10}

\bibitem{4}
Erik Cambria, Dipankar Das, Sivaji Bandyopadhyay, and Antonio Feraco.
\newblock Affective computing and sentiment analysis.
\newblock In {\em A Practical Guide to Sentiment Analysis}, pages 1--10. Springer, 2017.

\bibitem{25}
Shiming Chen, Wenjin Hou, Ziming Hong, Xiaohan Ding, Yibing Song, Xinge You, Tongliang Liu, and Kun Zhang.
\newblock Evolving semantic prototype improves generative zero-shot learning.
\newblock In {\em International Conference on Machine Learning}, pages 4611--4622. PMLR, 2023.

\bibitem{COVAREP30}
Gilles Degottex, John Kane, Thomas Drugman, Tuomo Raitio, and Stefan Scherer.
\newblock Covarep---a collaborative voice analysis repository for speech technologies.
\newblock In {\em 2014 IEEE International Conference on Acoustics, Speech and Signal Processing (ICASSP)}, pages 960--964. IEEE, 2014.

\bibitem{BERT32}
Jacob Devlin, Ming-Wei Chang, Kenton Lee, and Kristina Toutanova.
\newblock Bert: Pre-training of deep bidirectional transformers for language understanding.
\newblock In {\em Proceedings of the 2019 Conference of the North American Chapter of the Association for Computational Linguistics: Human Language Technologies, Volume 1 (Long and Short Papers)}, pages 4171--4186, 2019.

\bibitem{GAN19}
Ian~J. Goodfellow, Jean Pouget-Abadie, Mehdi Mirza, Bing Xu, David Warde-Farley, Sherjil Ozair, Aaron Courville, and Yoshua Bengio.
\newblock Generative adversarial nets.
\newblock In {\em Advances in Neural Information Processing Systems}, volume~27, 2014.

\bibitem{DDPM21}
Jonathan Ho, Ajay Jain, and Pieter Abbeel.
\newblock Denoising diffusion probabilistic models.
\newblock In {\em Advances in Neural Information Processing Systems}, volume~33, pages 6840--6851, 2020.

\bibitem{facet31}
{iMotions}.
\newblock Facial expression analysis.
\newblock \url{https://imotions.com/products/imotions-lab/modules/fea-facial-expression-analysis/}, 2017.
\newblock Accessed: 2025-05-03.

\bibitem{2}
Elsa~A. Kirchner, Stephen~H. Fairclough, and Frank Kirchner.
\newblock Embedded multimodal interfaces in robotics: Applications, future trends, and societal implications.
\newblock In {\em The Handbook of Multimodal-Multisensor Interfaces: Language Processing, Software, Commercialization, and Emerging Directions - Volume 3}, pages 523--576. ACM, 2019.

\bibitem{3}
Michelle~A. Lee, Yuke Zhu, Krishnan Srinivasan, Parth Shah, Silvio Savarese, Li~Fei-Fei, Animesh Garg, and Jeannette Bohg.
\newblock Making sense of vision and touch: Self-supervised learning of multimodal representations for contact-rich tasks.
\newblock In {\em 2019 International Conference on Robotics and Automation (ICRA)}, pages 8943--8950. IEEE, 2019.

\bibitem{GCNet10}
Zheng Lian, Lan Chen, Licai Sun, Bin Liu, and Jianhua Tao.
\newblock Gcnet: Graph completion network for incomplete multimodal learning in conversation.
\newblock {\em IEEE Transactions on Pattern Analysis and Machine Intelligence}, 2023.

\bibitem{MTMSA12}
Zhizhong Liu, Bin Zhou, Dianhui Chu, Yuhang Sun, and Lingqiang Meng.
\newblock Modality translation-based multimodal sentiment analysis under uncertain missing modalities.
\newblock {\em Information Fusion}, 101:101973, 2024.

\bibitem{Mai33}
Sijie Mai, Haifeng Hu, and Songlong Xing.
\newblock Modality to modality translation: An adversarial representation learning and graph fusion network for multimodal fusion.
\newblock {\em Proceedings of the AAAI Conference on Artificial Intelligence}, 34(01):164--172, 2020.

\bibitem{26}
Nikiforos Mimikos-Stamatopoulos, Benjamin Zhang, and Markos Katsoulakis.
\newblock Score-based generative models are provably robust: An uncertainty quantification perspective.
\newblock {\em Advances in Neural Information Processing Systems}, 37:63154--63183, 2024.

\bibitem{Peebles34}
William Peebles and Saining Xie.
\newblock Scalable diffusion models with transformers.
\newblock In {\em Proceedings of the IEEE/CVF International Conference on Computer Vision}, pages 4195--4205. IEEE, 2023.

\bibitem{5}
Soujanya Poria, Erik Cambria, Rajiv Bajpai, and Amir Hussain.
\newblock A review of affective computing: From unimodal analysis to multimodal fusion.
\newblock {\em Information Fusion}, 37:98--125, 2017.

\bibitem{7}
Soujanya Poria, Devamanyu Hazarika, Navonil Majumder, and Rada Mihalcea.
\newblock Beneath the tip of the iceberg: Current challenges and new directions in sentiment analysis research.
\newblock {\em IEEE Transactions on Affective Computing}, 2020.

\bibitem{LDM18}
Robin Rombach, Andreas Blattmann, Dominik Lorenz, Patrick Esser, and Bj{\"o}rn Ommer.
\newblock High-resolution image synthesis with latent diffusion models.
\newblock In {\em Proceedings of the IEEE/CVF Conference on Computer Vision and Pattern Recognition}, pages 10684--10695, 2022.

\bibitem{24}
Murat Sensoy, Lance Kaplan, Federico Cerutti, and Maryam Saleki.
\newblock Uncertainty-aware deep classifiers using generative models.
\newblock {\em Proceedings of the AAAI Conference on Artificial Intelligence}, 34(04):5620--5627, 2020.

\bibitem{6}
Mohammad Soleymani, Daniele Garcia, Brendan Jou, Bj{\"o}rn Schuller, Shih-Fu Chang, and Maja Pantic.
\newblock A survey of multimodal sentiment analysis.
\newblock {\em Image and Vision Computing}, 65:3--14, 2017.

\bibitem{8}
Kushal Somandepalli, Tanaya Guha, Victor~R. Martinez, Nithin Kumar, Hanchuan Adam, and Shrikanth Narayanan.
\newblock Computational media intelligence: Human-centered machine analysis of media.
\newblock {\em Proceedings of the IEEE}, 109(5):891--910, 2021.

\bibitem{9}
Leonhard Stappen, Alice Baird, Lukas Schumann, and Stefan Bjorn.
\newblock The multimodal sentiment analysis in car reviews (muse-car) dataset: Collection, insights and improvements.
\newblock {\em IEEE Transactions on Affective Computing}, 2021.

\bibitem{11}
Hao Sun, Jiaqing Liu, Yen-Wei Chen, and Lanfen Lin.
\newblock Modality-invariant temporal representation learning for multimodal sentiment classification.
\newblock {\em Information Fusion}, 91:504--514, 2023.

\bibitem{SMCMSA16}
Yuhang Sun, Zhizhong Liu, Quan~Z. Sheng, Dianhui Chu, Jian Yu, and Hongxiang Sun.
\newblock Similar modality completion-based multimodal sentiment analysis under uncertain missing modalities.
\newblock {\em Information Fusion}, 110:102454, 2024.

\bibitem{27}
Yao-Hung~Hubert Tsai, Shaojie Bai, Paul~Pu Liang, J.~Zico Kolter, Louis-Philippe Morency, and Ruslan Salakhutdinov.
\newblock Multimodal transformer for unaligned multimodal language sequences.
\newblock In {\em Proceedings of the 2019 Conference of the Association for Computational Linguistics (ACL)}, volume 2019, page 6558. NIH Public Access, 2019.

\bibitem{DiCMoR17}
Yuan Wang, Zhen Cui, and Yong Li.
\newblock Distribution-consistent modal recovering for incomplete multimodal learning.
\newblock In {\em Proceedings of the IEEE/CVF International Conference on Computer Vision}, pages 22025--22034, 2023.

\bibitem{IMDer13}
Yuan Wang, Yong Li, and Zhen Cui.
\newblock Incomplete multimodality-diffused emotion recognition.
\newblock {\em Advances in Neural Information Processing Systems}, 36, 2024.

\bibitem{NIAT20}
Ziqi Yuan, Yihe Liu, Hua Xu, and Kai Gao.
\newblock Noise imitation based adversarial training for robust multimodal sentiment analysis.
\newblock {\em IEEE Transactions on Multimedia}, 26:529--539, 2023.

\bibitem{MOSI28}
Amir Zadeh, Rowan Zellers, Eli Pincus, and Louis-Philippe Morency.
\newblock Multimodal sentiment intensity analysis in videos: Facial gestures and verbal messages.
\newblock {\em IEEE Intelligent Systems}, 31(6):82--88, 2016.

\bibitem{MOSEI29}
AmirAli~Bagher Zadeh, Paul~Pu Liang, Soujanya Poria, Erik Cambria, and Louis-Philippe Morency.
\newblock Multimodal language analysis in the wild: Cmu-mosei dataset and interpretable dynamic fusion graph.
\newblock In {\em Proceedings of the 56th Annual Meeting of the Association for Computational Linguistics (Volume 1: Long Papers)}, pages 2236--2246, 2018.

\bibitem{TATE15}
Jiandian Zeng, Tianyi Liu, and Jiantao Zhou.
\newblock Tag-assisted multimodal sentiment analysis under uncertain missing modalities.
\newblock In {\em Proceedings of the 45th International ACM SIGIR Conference on Research and Development in Information Retrieval}, pages 1545--1554, 2022.

\bibitem{MMIN14}
Jinming Zhao, Ruichen Li, and Qin Jin.
\newblock Missing modality imagination network for emotion recognition with uncertain missing modalities.
\newblock In {\em Proceedings of the 59th Annual Meeting of the Association for Computational Linguistics and the 11th International Joint Conference on Natural Language Processing (Volume 1: Long Papers)}, pages 2608--2618, 2021.

\end{thebibliography}

% \begin{thebibliography}{99}

  % 确保此处文件名与主文件一致（如 main.bbl）
% \end{thebibliography}

\end{document}